\documentclass[11pt,a4paper]{article}
\usepackage[hyperref]{acl2018}
\usepackage{times}
\usepackage{latexsym}
\usepackage{hyperref}
\usepackage{url}
\usepackage{graphicx}
\usepackage{tabularx}
\usepackage{multirow}
\usepackage{booktabs}
\usepackage{subcaption}
\usepackage{float}  
\usepackage{amsmath}
\usepackage{amsfonts,amssymb}
\usepackage[T1]{fontenc}
\usepackage[utf8]{inputenc}
\usepackage{tikz-dependency}
\usepackage[export]{adjustbox}
\newcommand{\benchmark}[1]{{\it #1}}

\usepackage{nicefrac}

\usepackage{caption}
\definecolor{darkred}{HTML}{d53e4f}
\definecolor{myblue}{HTML}{3288bd}
\newcommand{\mathbold}[1]{\ensuremath{\boldsymbol{\mathbf{#1}}}}
\newcommand{\g}{\,|\,}

\newcommand{\T}{\mathsf{T}}

\newcommand{\R}{\mathbb{R}}
\newcommand{\mbh}{\mathbold{h}}
\newcommand{\mby}{\mathbold{y}}
\newcommand{\mbc}{\mathbold{c}}
\newcommand{\mbd}{\mathbold{d}}
\newcommand{\mbu}{\mathbold{u}}
\newcommand{\mbs}{\mathbold{s}}

\newcommand{\mbx}{\mathbold{x}}
\newcommand{\mbz}{\mathbold{z}}
\newcommand{\mbW}{\mathbold{W}}

\newcommand{\mbL}{\mathbold{L}}
\newcommand{\mbM}{\mathbold{M}}
\newcommand{\mbS}{\mathbold{S}}

\newcommand{\mbtheta}{\mathbold{\theta}}
\newcommand{\mbbeta}{\mathbold{\beta}}
\newcommand{\mbalpha}{\mathbold{\alpha}}
\newcommand{\mbphi}{\mathbold{\phi}}
\newcommand{\mbgamma}{\mathbold{\gamma}}

\newcommand{\mi}{$\mbox{}^{\vartriangle}$\hspace*{-.0em}\mbox{}}
\newcommand{\si}{$\mbox{}^\blacktriangle$\hspace*{-.0em}\mbox{}}

\usepackage[colorinlistoftodos,prependcaption,textsize=small]{todonotes}

\aclfinalcopy 

\title{Inducing Grammars with and for Neural Machine Translation}

\author{Ke Tran \\
Informatics Institute \\
University of Amsterdam \\
\texttt{ketranmanh@gmail.com} \\
\And
Yonatan Bisk \\
Department of Computer Science \\
University of Washington \\
\texttt{ybisk@cs.washington.com}
}

\date{}

\begin{document}
\maketitle
\begin{abstract}
  Machine translation systems require semantic knowledge and grammatical understanding.  Neural machine translation (NMT) systems often assume this information is captured by an attention mechanism and a decoder that ensures fluency.  Recent work has shown that incorporating explicit syntax alleviates the burden of modeling both types of knowledge.  However, requiring parses is expensive and does not explore the question of what syntax a model needs during translation.  To address both of these issues we introduce a model that simultaneously translates while inducing dependency trees.  In this way, we leverage the benefits of structure while investigating what syntax NMT must induce to maximize performance.  We show that our dependency trees are 1. language pair dependent and 2. improve translation quality.
\end{abstract}

\section{Motivation}
\label{sec:intro}
Language has syntactic structure and translation models need to understand grammatical dependencies to resolve the semantics of a sentence and preserve agreement ({\em e.g.}, number, gender, etc).  Many current approaches to MT have been able to avoid explicitly providing structural information by relying on advances in sequence to sequence (seq2seq) models.  The most famous advances include attention mechanisms \citep{bahdanau:2015} and gating in Long Short-Term Memory (LSTM) cells \citep{Hochreiter:1997}.

In this work we aim to benefit from syntactic structure, without providing it to the model, and to disentangle the semantic and syntactic components of translation, by introducing a gating mechanism which controls when syntax should be used.

\begin{figure}[t]
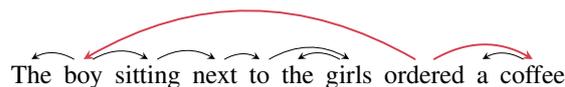

\centering
\scalebox{0.9}{
\begin{dependency}[arc edge, arc angle=30, text only label, label style={above,scale=.5}]
  \begin{deptext}
  The \& boy \& sitting \& next \& to \& the \& girls \& ordered \& a \& coffee\\
  \end{deptext}
  \depedge[edge style={darkred,thick}]{8}{2}{}
  \depedge[edge style={darkred,thick}]{8}{10}{}
  \depedge{10}{9}{}
  \depedge{2}{1}{}
  \depedge{2}{3}{}
  \depedge{3}{4}{}
  \depedge{4}{5}{}
  \depedge{5}{7}{}
  \depedge{7}{6}{}
\end{dependency}}
\caption{Our model aims to capture both: \\
\textbf{syntactic} (verb \textit{ordered} \textcolor{red}{$\rightarrow$} subj/obj \textit{boy, coffee})\\
\textbf{alignment} (noun \textit{girls} $\rightarrow$ determiner \textit{the}) attention.}
\label{fig:example}
\end{figure}

Consider the process of translating the sentence ``\textit{The boy sitting next to the girls ordered a coffee.}'' (Figure \ref{fig:example}) from English to German.
In German, translating \emph{ordered}, requires knowledge of its subject \textit{boy} to correctly predict the verb's number \textit{bestellte} instead of \textit{bestellten}. 
This is a case where syntactic agreement requires long-distance information.
On the other hand, \emph{next} can be translated in isolation.
The model should uncover these relationships and decide when and which aspects of syntax are necessary.
While in principle decoders can utilize previously predicted words ({\em e.g.}, the translation of \emph{boy}) to reason about subject-verb agreement, in practice LSTMs still struggle with long-distance dependencies. Moreover, \citet{Belinkov17} showed that using attention reduces the decoder's capacity to learn target side syntax.

In addition to demonstrating improvements in translation quality, we are also interested in analyzing the predicted dependency trees discovered by our models. Recent work has begun analyzing task-specific latent trees \citep{Williams2017}.
We present the first results on learning latent trees with a joint syntactic-semantic objective.  We do this in the service of machine translation which inherently requires access to both aspects of a sentence.  Further, our results indicate that language pairs with rich morphology require and therefore induce more complex syntactic structure.

Our use of a structured self attention encoder (\S\ref{sec:models}) that predicts a non-projective dependency tree over the source sentence provides a soft structured representation of the source sentence that can then be transferred to the decoder, which  alleviates the burden of capturing target syntax on the target side.

We will show that the quality of the induced trees depends on the choice of the target language (\S\ref{sec:grammar_induction}). Moreover, the gating mechanism will allow us to examine which contexts require source side syntax.

In summary, in this work:
\begin{itemize}
  \item We propose a new NMT model that discovers latent structures for encoding and when to use them, while achieving significant improvements in BLEU scores over a strong baseline.
  \item We perform an in-depth analysis of the induced structures and investigate where the target decoder decides syntax is required.
\end{itemize}

\section{Related Work}
Recent work has begun investigating what syntax seq2seq models capture \citep{Linzen16}, but this is evaluated via downstream tasks designed to test the model's abilities and not its representation.

Simultaneously, recent research in neural machine translation (NMT) has shown the benefit of modeling syntax explicitly \citep{aharoni-goldberg:2017:Short,Bastings17,Li17,Eriguchi17} rather than assuming the model will automatically discover and encode it.

\citet{bradbury-socher:2017:StructPred} presented an encoder-decoder architecture based on RNNG \cite{dyer-EtAl:2016:N16-1}. However, their preliminary work was not scaled to a large MT dataset and omits analysis of the induced trees.

Unlike the previous work on source side latent graph parsing \citep{Hashimoto17}, our structured self attention encoder allows us to extract a dependency tree in a principled manner. Therefore, learning the internal representation of our model is related to work done in unsupervised grammar induction \cite{Klein:2004aa,Spitkovsky:2011ab} except that by focusing on translation we require both syntactic and semantic knowledge.

In this work, we attempt to contribute to both modeling syntax and investigating a more interpretable interface for testing the syntactic content of a new seq2seq models' internal representation.
\section{Neural Machine Translation}
\label{sec:nmt}
Given a training pair of source and target sentences $(\mbx, \mby)$ of length $n$ and $m$ respectively, neural machine translation is a conditional probabilistic model $p(\mby \g \mbx)$ implemented using neural networks
\begin{align}\nonumber
  \log p(\mby \g \mbx;\;\mbtheta) &= \sum_{j=1}^m \log p(\mby_j\g\mby_{i < j}, \mbx;\; \mbtheta)
\end{align}
where $\mbtheta$ is the model's parameters. We will omit the parameters $\mbtheta$ herein for readability.

The NMT system used in this work is a seq2seq model that consists of a bidirectional LSTM encoder and an LSTM decoder coupled with an attention mechanism \citep{bahdanau:2015,luong-pham-manning:2015:EMNLP}. Our system is based on a PyTorch implementation\footnote{\url{http://opennmt.net/OpenNMT-py/}} of OpenNMT \citep{klein:2017}.
Let $\{\mbs_i \in\R^d \}_{i=1}^n$ be the output of the encoder
\begin{align}
  \mbS &= \textrm{BiLSTM}(\mbx)
\end{align}
Here we use $\mbS = [\mbs_1;\dots;\mbs_n] \in \R^{d\times n}$ as a concatenation of $\{\mbs_i\}$.
The decoder is composed of stacked LSTMs with input-feeding.
Specifically, the inputs of the decoder at time step $t$ are a concatenation of the embedding of the previous generated word $\mby_{t-1}$ and a vector $\mbu_{t-1}$:
\begin{align}\label{eq:u}
  \mbu_{t-1} &= g(\mbh_{t-1}, \mbc_{t-1})
\end{align}
where $g$ is a one layer feed-forward network, $\mbh_{t-1}$ is the output of the LSTM decoder, and $\mbc_{t-1}$ is a context vector computed by an attention mechanism
\begin{align}
  \mbalpha_{t-1} &= \textrm{softmax}(\mbh_{t-1}^\T\mbW_a\mbS) \label{eq:attn_weight}\\ 
  \mbc_{t-1} &= \mbS\mbalpha_{t-1}^\T \label{eq:context}
\end{align}
where $\mbW_a \in \R^{d\times d}$ is a trainable parameter.

Finally a single layer feed-forward network $f$ takes $\mbu_t$ as input and returns a multinomial distribution over all the target words: $y_t \sim f(\mbu_t)$

\section{Syntactic Attention Model}
\label{sec:models}
\begin{figure*}[htbp]
  \centering
  \begin{subfigure}[t]{0.23\textwidth}
      \includegraphics[width=\textwidth]{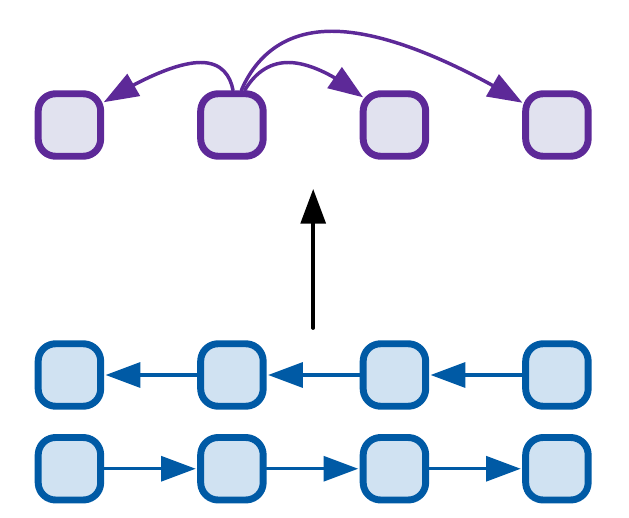}
      \caption{Structured Self Attention Encoder: the first layer is a standard BiLSTM, the top layer is a syntactic attention network.}
      \label{fig:syntax_encoder}
  \end{subfigure}
  \qquad
  \begin{subfigure}[t]{0.23\textwidth}
      \includegraphics[width=\textwidth]{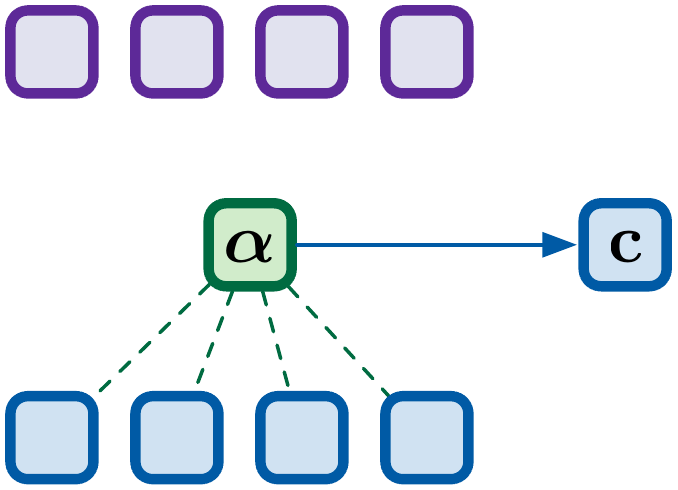}
      \caption{Compute the context vector (blue) as in a standard NMT model. The attention weights $\mbalpha$ are in green.}
      \label{fig:context_vector}
  \end{subfigure}
  \qquad
  \begin{subfigure}[t]{0.23\textwidth}
      \includegraphics[width=\textwidth]{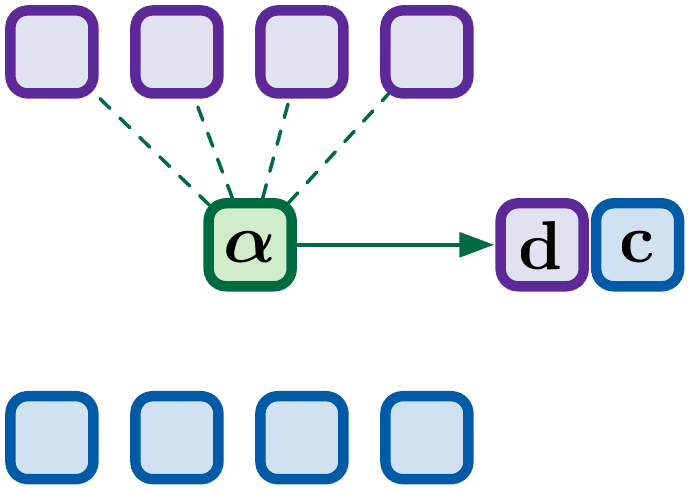}
      \caption{Use the attention weights $\mbalpha$, as computed in the previous step, to calculate syntactic vector (purple).}
      \label{fig:syntactic_vector}
  \end{subfigure}
  \caption{A visual representation of our proposed mechanism for shared attention.}\label{fig:models}
\end{figure*}

We propose a syntactic attention model\footnote{\url{https://github.com/ketranm/sa-nmt}} (Figure~\ref{fig:models}) that differs from standard NMT in two crucial aspects. First, our encoder outputs two sets of annotations: content annotations $\mbS$ and syntactic annotations $\mbM$ (Figure~\ref{fig:syntax_encoder}). The content annotations are the outputs of a standard BiLSTM while the syntactic annotations are produced by a head word selection layer (\S\ref{ssec:structured_attention}).
The syntactic annotations $\mbM$ capture syntactic dependencies amongst the source words and enable syntactic transfer from the source to the target.
Second, we incorporate the source side syntax into our model by modifying the standard attention (from target to source) in NMT such that it attends to both $\mbS$ and $\mbM$ through a \textit{shared attention} layer. The shared attention layer biases our model toward capturing source side dependency. It produces a dependency context $\mbd$ (Figure~\ref{fig:syntactic_vector}) in addition to the standard context vector $\mbc$ (Figure~\ref{fig:context_vector}) at each time step.
Motivated by the example in Figure~\ref{fig:example} that some words can be translated without resolving their syntactic roles in the source sentence, we include a gating mechanism that allows the decoder to decide the amount of syntax needed when it generates the next word.
Next, we describe the head word selection layer and how source side syntax is incorporated into our model.

\subsection{Head Word Selection}
\label{ssec:structured_attention}
The head word selection layer learns to select a \emph{soft} head word for each source word.
This layer transforms $\mbS$ into a matrix $\mbM$ that encodes implicit dependency structure of $\mbx$ using \emph{structured self attention}. First we apply three trainable weight matrices $\mbW_q, \mbW_k, \mbW_v \in \R^{d\times d}$ to map $\mbS$ to query, key, and value matrices $\mbS_q=\mbW_q\mbS$, $\mbS_k=\mbW_k\mbS$, $\mbS_v=\mbW_v\mbS \in\R^{d\times n}$ respectively.
Then we compute the structured self attention probabilities $\mbbeta\in\R^{n\times n}$ via a function \emph{sattn}: $\mbbeta  = \textrm{sattn}(\nicefrac{\mbS_q^\T \mbS_k}{\sqrt{d}})$. Finally the syntactic context $\mbM$ is computed as $\mbM = \mbS_v\mbbeta$.

Here $n$ is the length of the source sentence, so $\mbbeta$ captures all pairwise word dependencies.  Each cell $\mbbeta_{i,j}$ of the attention matrix $\mbbeta$ is the posterior probability $p(x_i=\text{head}(x_j)\g \mbx)$.
The structured self attention function \emph{sattn} is inspired by the work of \citep{Kim17} but differs in two important ways. First we model \textit{non-projective dependency trees}. Second, we utilize the Kirchhoff's Matrix-Tree Theorem \citep{Tutte84}  instead of the sum-product algorithm presented in \citep{Kim17} for fast evaluation of the attention probabilities. We note that \cite{Liu2017} were first to propose using the Matrix-Tree Theorem for evaluating the marginals in end to end training of neural networks.
Their work, however, focuses on the task of natural language inference \citep{Bowman:snli15} and document classification which arguably require less syntactic knowledge than machine translation.
Additionally, we will evaluate our structured self attention on datasets that are up to 20 times larger than the datasets studied in previous work.

Let $\mbz \in \{0, 1\}^{n\times n}$ be an adjacency matrix encoding a source's dependency tree. Let $\mbphi = \nicefrac{\mbS_q^\T\mbS_k}{\sqrt{d}} \in\R^{n\times n}$ be a scoring matrix such that cell $\mbphi_{i,j}$ scores how likely word $x_i$ is to be the head of word $x_j$.
The probability of a dependency tree $\mbz$ is therefore given by
\begin{align}
  p(\mbz\g\mbx; \mbphi) &= \frac{\exp\left(\sum_{i,j} \mbz_{i,j} \; \mbphi_{i,j}\right)}{Z(\mbphi)}
\end{align}
where $Z(\mbphi)$ is the partition function.

In the head selection model, we are interested in the marginal $p(\mbz_{i,j}= 1 \g \mbx; \mbphi)$
\begin{align}
 \mbbeta_{i,j}  &=  p(\mbz_{i,j}= 1 \g \mbx; \mbphi) &= \sum_{\mbz\;:\; \mbz_{i,j}=1} p(\mbz\g\mbx; \mbphi) \nonumber
\end{align}
We use the framework presented by \citet{koo2007} to compute the marginal of non-projective dependency structures. \citet{koo2007} use the Kirchhoff's Matrix-Tree Theorem \citep{Tutte84} to compute $p(\mbz_{i,j}= 1 \g \mbx; \mbphi)$ by first defining the Laplacian matrix $\mbL \in \R^{n\times n}$ as follows:
\begin{align}
  \mbL_{i,j} (\mbphi) &=
  \begin{cases}
    \sum\limits_{\substack{k=1\\ k\neq j}}^n \exp(\mbphi_{k,j}) & \text{if } i=j\\
    -\exp(\mbphi_{i,j})             & \text{otherwise}
  \end{cases}
\end{align}
Now we construct a matrix $\hat\mbL$ that accounts for root selection
\begin{align}
  \hat\mbL_{i,j} (\mbphi) &=
  \begin{cases}
    \exp(\mbphi_{j,j}) & \text{if } i=1\\
    \mbL_{i,j} (\mbphi)   & \text{if } i>1
  \end{cases}
\end{align}
The marginals in $\mbbeta$ are then
\begin{align}
\mbbeta_{i,j} &= (1-\delta_{1,j})\exp(\mbphi_{i,j})\left[\hat\mbL^{-1}(\mbphi)\right]_{j,j}  \nonumber\\
              &- (1-\delta_{i,1})\exp(\mbphi_{i,j})\left[\hat\mbL^{-1}(\mbphi)\right]_{j,i}
\end{align}
where $\delta_{i,j}$ is the Kronecker delta.
For the root node, the marginals are given by
\begin{align}
  \mbbeta_{k,k} = \exp(\mbphi_{k,k})\left[\hat\mbL^{-1}(\mbphi)\right]_{k, 1}
\end{align}

The computation of the marginals is fully differentiable, thus we can train the model in an end-to-end fashion by maximizing the conditional likelihood of the translation.

\subsection{Incorporating Syntactic Context}
\label{ssec:gate}
Having set the annotations $\mbS$ and $\mbM$ with the encoder, the LSTM decoder can utilize this information at every generation step by means of attention. At time step $t$, we first compute standard attention weights $\mbalpha_{t-1}$ and context vector $\mbc_{t-1}$ as in Equations (\ref{eq:attn_weight}) and (\ref{eq:context}). We then compute a weighted syntactic vector:
\begin{align}\label{eq:shared_att}
  \mbd_{t-1} &= \mbM\mbalpha_{t-1}^\T
\end{align}
Note that the syntactic vector $\mbd_{t-1}$ and the context vector $\mbc_{t-1}$ share the same attention weights $\mbalpha_{t-1}$. The main idea behind sharing attention weights (Figure~\ref{fig:syntactic_vector}) is that if the model attends to a particular source word $x_i$ when generating the next target word, we also want the model to attend to the head word of $x_i$.
We share the attention weights $\mbalpha_{t-1}$ because we expect that, if the model picks a source word $x_i$ to translate with the highest probability $\mbalpha_{t-1}[i]$, the contribution of $x_i$'s head in the syntactic vector $\mbd_{t-1}$ should also be highest.
\begin{figure}[ht]
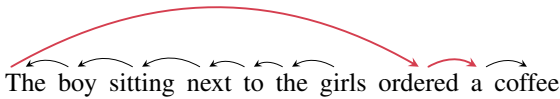

\centering
\scalebox{0.9}{
\begin{dependency}[arc edge, arc angle=30, text only label, label style={above,scale=.5}]
  \begin{deptext}
  The \& boy \& sitting \& next \& to \& the \& girls \& ordered \& a \& coffee\\
  \end{deptext}
  \depedge{7}{6}{}
  \depedge{2}{1}{}
  \depedge[edge start x offset=-10pt,edge style={darkred,thick}]{1}{8}{}
  \depedge{3}{2}{}
  \depedge{4}{3}{}
  \depedge{5}{4}{}
  \depedge{6}{5}{}
  \depedge[edge style={darkred,thick}]{8}{9}{}
  \depedge{9}{10}{}
\end{dependency}}
\caption{A latent tree learned by our model.}
\label{fig:ex_model}
\end{figure}
Figure~\ref{fig:ex_model} shows the latent tree learned by our translation objective. Unlike the gold tree provided in Figure~\ref{fig:example}, the model decided that ``\textit{the boy}'' is the head of ``\textit{ordered}''. This is common in our model because the BiLSTM context means that a given word's representation is actually a summary of its local context/constituent.

It is not always useful or necessary to access the syntactic context $\mbd_{t-1}$ at every time step $t$. Ideally, we should let the model decide whether it needs to use this information or not. For example, the model might decide to only use syntax when it needs to resolve long distance dependencies on the source side. To control the amount of source side syntactic information, we introduce a gating mechanism:
\begin{align}
\hat\mbd_{t-1} = \mbd_{t-1} \,\odot\, \sigma(\mbW_g\mbh_{t-1})
\end{align}
The vector $\mbu_{t-1}$ from Eq.~(\ref{eq:u}) now becomes
\begin{align}
  \mbu_{t-1} &= g(\mbh_{t-1}, \mbc_{t-1}, \hat\mbd_{t-1})
\end{align}

Another approach to incorporating syntactic annotations $\mbM$ in the decoder is to use a separate attention layer to compute the syntactic vector $\mbd_{t-1}$ at time step $t$:
\begin{align}
  \mbgamma_{t-1} &= \textrm{softmax}(\mbh_{t-1}^\T\mbW_m\mbM) \\
  \mbd_{t-1} &= \mbM\mbgamma_{t-1}^\T
\end{align}

We will provide a comparison to this approach in our results.

\subsection{Hard Attention over Tree Structures}
Finally, to simulate the scenario where the model has access to a dependency tree given by an external parser we report results with hard attention. Forcing the model to make hard decisions during training mirrors the extraction and conditioning on a dependency tree (\S\ref{ssec:extract_tree}). We expect
this technique will improve the performance on grammar induction, despite making translation lossy.
A similar observation has been reported in \citep{Hashimoto17} which showed that translation performance degraded below their baseline when they provided dependency trees to the encoder.

Recall the marginal $\mbbeta_{i,j}$ gives us the probability that word $x_i$ is the head of word $x_j$. We convert these soft weights to hard ones $\bar\mbbeta$ by
\begin{align}\label{eq:greedy}
  \bar\mbbeta_{k,j} &=
  \begin{cases}
    1 &\text{if } k=\arg\max_{i}\mbbeta_{i,j} \\
    0 &\text{otherwise}
  \end{cases}
\end{align}
We train this model using the straight-through estimator \citep{bengio2013estimating}. In this setup, each word has a parent but there is no guarantee that the structure given by hard attention will result in a tree ({\em i.e.}, it may contain cycle). A more principled way to enforce a tree structure is to decode the best tree $\mathcal{T}$ using the maximum spanning tree algorithm \citep{chu-liu-1965,Edmonds} and to set $\bar\mbbeta_{k,j}=1$ if the edge $(x_k\to x_j) \in \mathcal{T}$. Maximum spanning tree decoding can be prohibitively slow as the Chu-Liu-Edmonds algorithm is not GPU friendly.
We therefore greedily pick a parent word for each word $x_j$ in the sentence using Eq.~(\ref{eq:greedy}). This is actually a principled simplification as greedily assigning a parent for each word is the first step in Chu-Liu-Edmonds algorithm.

\section{Experiments}
\label{sec:experiment}
Next we will discuss our experimental setup and report results for English$\leftrightarrow$German (En$\leftrightarrow$De), English$\leftrightarrow$Russian (En$\leftrightarrow$Ru), and Russian$\rightarrow$Arabic (Ru$\rightarrow$Ar) translation models.
\subsection{Data}
We use the WMT17 \citep{bojar-EtAl:2017:WMT1} data in our experiments. Table~\ref{tb:data} shows the statistics of the data. For En$\leftrightarrow$De, we use a concatenation of Europarl, Common Crawl, Rapid corpus of EU press releases, and News Commentary v12. We use \emph{newstest2015} for development and \emph{newstest2016}, \emph{newstest2017} for testing.
For En$\leftrightarrow$Ru, we use Common Crawl, News Commentary v12, and Yandex Corpus. The development data comes from \emph{newstest2016} and \emph{newstest2017} is reserved for testing.
For Ru$\rightarrow$Ar, we use the data from the six-way sentence-aligned subcorpus of the United Nations Parallel Corpus v1.0 \cite{Ziemski:16}. The corpus also contains the official development and test data.
\begin{table}[ht]
  \begin{footnotesize}
  \centering
  \begin{tabular}{@{} l r r c c@{}}
  \toprule
    & Train & Valid & Test & Vocabulary\\
  \midrule
  En$\leftrightarrow$De & 5.9M & 2,169 & 2,999 / 3,004 & 36,251 / 35,913\\
  En$\leftrightarrow$Ru & 2.1M & 2,998 & 3,001 & 34,872 / 34,989\\
  Ru$\rightarrow$Ar    & 11.1M & 4,000 & 4,000 & 32,735 / 32,955\\
  \bottomrule
  \end{tabular}
  \end{footnotesize}
  \vspace{-5pt}
  \caption{Statistics of the data.}
  \label{tb:data}
\end{table}
Our language pairs were chosen to compare results across and between morphologically rich and poor languages.  This will prove particularly interesting in our grammar induction results where different pairs must preserve different amounts of syntactic agreement information.

We use BPE \citep{sennrich-haddow-birch:2016:P16-12} with 32,000 merge operations. We run BPE for each language instead of using BPE for the concatenation of both source and target languages.
\subsection{Baselines}
Our baseline is an NMT model with input-feeding (\S\ref{sec:nmt}). As we will be making several modifications from the basic architecture in our proposed structured self attention NMT (SA-NMT), we will verify each choice in our architecture design empirically. First we validate the structured self attention module by comparing it to a self-attention module \citep{lin:2017,Vaswani17}. Self attention computes attention weights $\mbbeta$ simply as $\mbbeta=\textrm{softmax}(\mbphi)$. Since self-attention does not assume any hierarchical structure over the source sentence, we refer it as flat-attention NMT (FA-NMT).
Second, we validate the benefit of using two sets of annotations in the encoder. We combine the hidden states of the encoder $\mbh$ with syntactic context $\mbd$ to obtain a single set of annotation using the following equation:
\begin{align}
  \bar\mbs_i = \mbs_i + \sigma(\mbW_g\mbs_i) \odot \mbd_i
\end{align}
Here we first down-weight the syntactic context $\mbd_i$ before adding it to $\mbs_i$. The sigmoid function $\sigma(\mbW_g\mbs_i)$ decides the weight of the head word of $x_i$ based on whether translating $x_i$ needs additionally dependency information. We refer to this baseline as SA-NMT-1set. Note that in this baseline, there is only one attention layer from the target to the source $\bar\mbS=\{\bar\mbs_i\}_1^n$.

In all the models, we share the weights of target word embeddings and the output layer as suggested by \citet{Inan:2016} and \citet{press:2017}.

\subsection{Hyper-parameters and Training}
For all the models, we set the word embedding size to 1024, the number of LSTM layers to 2, and the dropout rate to 0.3. Parameters are initialized uniformly in $(-0.04, 0.04)$. We use the Adam optimizer  \citep{kingma:2014} with an initial learning rate of 0.001.
We evaluate our models on development data every 10,000 updates for De--En and Ru$\rightarrow$Ar, and 5,000 updates for Ru--En. If the validation perplexity increases, we decay the learning rate by 0.5. We stop training after decaying the learning rate five times as suggested by \citet{denkowski-neubig:2017:NMT}. The mini-batch size is 64 in Ru$\rightarrow$Ar experiments and 32 in the rest. Finally, we report BLEU scores computed using the standard \texttt{multi-bleu.perl} script.

In our experiments, the SA-NMT models are twice slower than the baseline NMT measuring by the number of target words generated per second.
\subsection{Translation Results}
\begin{table*}[ht]
  \centering
  \begin{footnotesize}
  \begin{tabular}{@{} l l l l l l l l l@{}}
  \toprule
  Model  & Shared & \multicolumn{2}{c}{De$\to$En} & Ru$\to$En  & \multicolumn{2}{c}{En$\to$De} & En$\to$Ru & Ru$\to$Ar\\
         &      & \benchmark{wmt16} & \benchmark{wmt17} &  \benchmark{wmt17} & \benchmark{wmt16} & \benchmark{wmt17} &  \benchmark{wmt17} & \benchmark{un-test}\\
  \midrule
  NMT & - & 33.16 &  28.94 & 30.17 & 29.92 &  23.44 & 26.41 & 37.04\\
  \midrule
  \multirow{ 2}{*}{FA-NMT} & yes & 33.55 & 29.43 & 30.22  &  30.09 & 24.03 & 26.91 & 37.41\\
   & no & 33.24 & 29.00 & 30.34 & 29.98 & 23.97 & 26.75 & 37.20\\
  \midrule
  SA-NMT-1set & - & 33.51 & 29.15 & 30.34 & \bf{30.29}$^{\dagger}$ & 24.12 & 26.96  & 37.34\\
  SA-NMT-hard & yes & 33.38 & 28.96 & 29.98 & 29.93 & 23.84 &  26.71 & 37.33\\
  \midrule
  \multirow{ 2}{*}{SA-NMT} & yes &  \bf{33.73}$^{\ddagger}$\mi & \bf{29.45}$^{\ddagger}$\si & \bf{30.41} &  30.22 & \bf{24.26}$^{\ddagger}$\mi & \bf{27.05}$^{\ddagger}$  & \bf{37.49}$^{\ddagger}$\mi \\
   & no & 33.18 & 29.19 & 30.15 & 30.17 & 23.94 & 27.01 & 37.22 \\
  \bottomrule
  \end{tabular}
  \end{footnotesize}
  \caption{Results for translating En$\leftrightarrow$De, En$\leftrightarrow$Ru, and Ru$\rightarrow$Ar.
  Statistical significances are marked as $^\dagger p<0.05$  and $^\ddagger p<0.01$ when compared against the baselines and \mi/\si\  when compared against the FA-NMT (no-shared). The results indicate the strength of our proposed \emph{shared-attention} for NMT.}
  \label{tb:bleus}
  \vspace{-5pt}
\end{table*}
Table~\ref{tb:bleus} shows the BLEU scores in our experiments.  We test statistical significance using bootstrap resampling \citep{riezler-maxwell:2005:MTSumm}. Statistical significances are marked as $^\dagger p<0.05$  and $^\ddagger p<0.01$ when compared against the baselines.
Additionally, we also report statistical significances \mi $p<0.05$ and \si $p<0.01$ when comparing against the FA-NMT models that have two separate attention layers from the decoder to the encoder. Overall, the SA-NMT (shared) model performs the best gaining more than 0.5 BLEU De$\to$En on \benchmark{wmt16}, up to 0.82 BLEU on En$\to$De \benchmark{wmt17} and 0.64 BLEU En$\to$Ru direction over a competitive NMT baseline.
The gain of the SA-NMT model on Ru$\to$Ar is small (0.45 BLEU) but significant.
The results show that structured self attention is useful when translating from English to languages that have long-distance dependencies and complex morphological agreements. We also see that the gain is marginal compared to self-attention models (FA-NMT-shared) and not significant. Within FA-NMT models, sharing attention is helpful. Our results also confirm the advantage of having two separate sets of annotations in the encoder when modeling syntax.
The hard structured self attention model (SA-NMT-hard) performs comparably to the baseline. While this is a somewhat expected result from the hard attention model, we will show in Section \ref{sec:grammar_induction} that the quality of induced trees from hard attention is often far better than those from soft attention.
\begin{table*}
\centering
\begin{footnotesize}
\begin{tabular}{lcc c ccc c ccc}
\toprule
     &    \multicolumn{2}{c}{FA} &  & \multicolumn{3}{c}{SA} &  & \multicolumn{3}{c}{Baseline}\\
     \cmidrule{2-3} \cmidrule{5-7}  \cmidrule{9-11}
     &  no-shared & shared  &  & no-shared & shared & hard & & L & R & Un\\
\midrule
EN ($\rightarrow$DE) & 17.0/25.2 & 27.6/41.3 &  & 23.6/33.7 & 27.8/42.6 & \textbf{31.7}/\textbf{45.6} & &  \multirow{ 2}{*}{34.0} & \multirow{ 2}{*}{\phantom{0}7.8} & \multirow{ 2}{*}{40.9} \\
EN ($\rightarrow$RU) & 35.2/48.5 & \textbf{36.5}/48.8 & & 12.8/25.5 & 33.1/\textbf{48.9} & 33.7/46.0 & & \\
\midrule
DE ($\rightarrow$EN) & 21.1/33.3 & 20.1/33.6 & & 12.8/22.5 & 21.5/38.0  & \textbf{26.3}/\textbf{40.7} & & 34.4 & \phantom{0}8.6 & 41.5\\
\midrule
RU ($\rightarrow$EN) & 19.2/33.2 & 20.4/34.9 & & 19.3/34.4 & \textbf{24.8}/\textbf{41.9} & 23.2/33.3 & & \multirow{ 2}{*}{32.9} & \multirow{ 2}{*}{15.2} & \multirow{ 2}{*}{47.3}\\
RU ($\rightarrow$AR) & 21.1/41.1 & 22.2/42.1 &  & 11.6/21.4 & 28.9/50.4 & \textbf{30.3}/\textbf{52.0} & & \\
\bottomrule
\end{tabular}
\end{footnotesize}
\caption{Directed and Undirected (DA/UA) model accuracy (without punctuation) compared to branching baselines: left (L), right (R) and undirected (Un). Our results show an intriguing effect of the target language on induction.  Note the accuracy discrepancy between translating RU to EN versus AR.}
\label{DAUA}
\vspace{-5pt}
\end{table*}

\section{Gate Activation Visualization}
\label{sec:vis}
\begin{figure}[htbp]
  \centering
  \includegraphics[max width=\columnwidth]{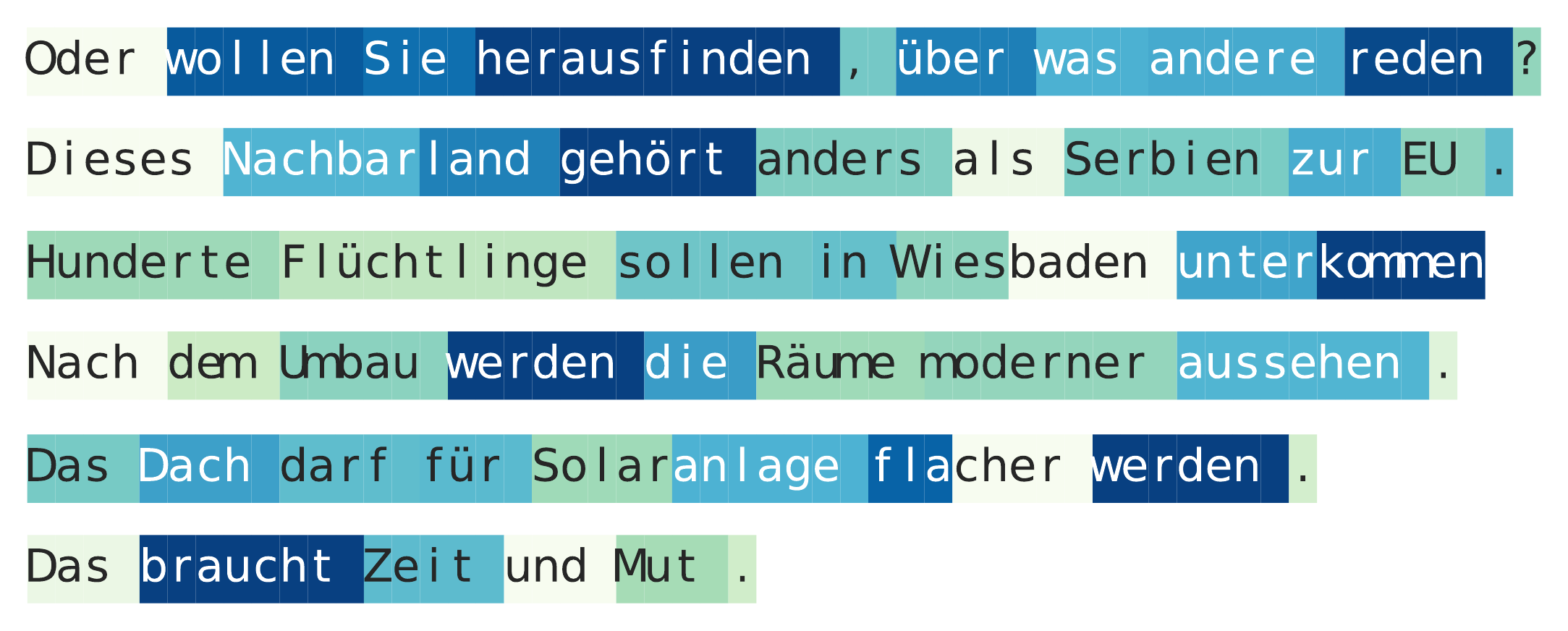}
  \vspace{-5pt}
  \caption{Visualization of gate norm. Darker means the model is using more syntactic information.}
  \vspace{-5pt}
  \label{fig:gatenorm}
\end{figure}

As mentioned earlier, our models allow us to ask the question: When does the target LSTM need to access source side syntax?  We investigate this by analyzing the gate activations of our best model, SA-NMT (shared). At time step $t$, when the model is about to predict the target word $y_t$, we compute the norm of the gate activations
\begin{align}
  z_t &= \lVert\sigma(\mbW_g\mbh_{t-1}) \rVert_2
\end{align}
The activation norm $z_t$ allows us to see how much syntactic information flows into the decoder. We observe that $z_t$ has its highest value when the decoder is about to generate a verb while it has its lowest value when the end of sentence token \verb|</s>| is predicted. Figure~\ref{fig:gatenorm} shows some examples of German target sentences. The darker colors represent higher activation norms.

It is clear that translating verbs requires structural information. We also see that after verbs, the gate activation norms are highest at nouns Zeit (\textit{time}), Mut (\textit{courage}), Dach (\textit{roof}) and then tail off as we move to function words which require less context to disambiguate. Below are the frequencies with which the highest activation norm in a sentence is applied to a given part-of-speech tag on \emph{newstest2016}.  We include the top 7 most common activations. We see that while nouns are often the most common tag in a sentence, syntax is disproportionately used for translating verbs.
\begin{figure}[ht]
\centering
\includegraphics[width=0.7\linewidth]{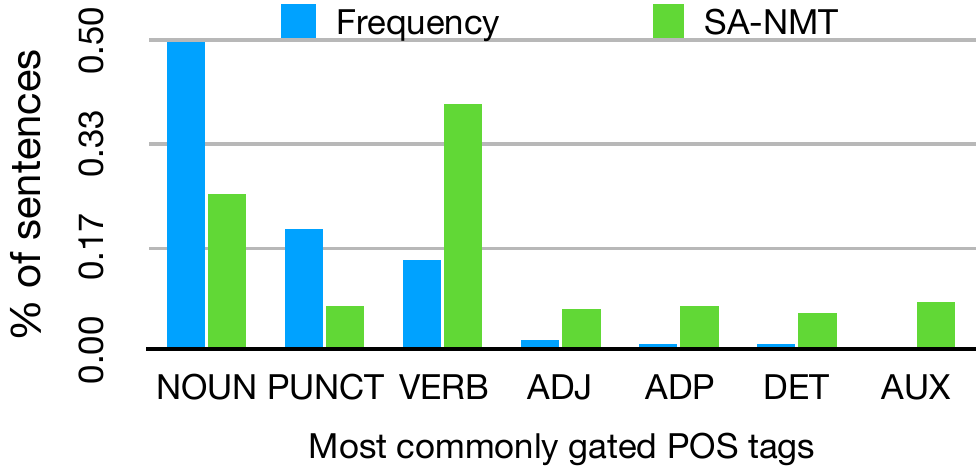}
\vspace{-5pt}
\end{figure}

\section{Grammar Induction}
\label{sec:grammar_induction}
NLP has long assumed hierarchical structured representations are important to understanding language.  In this work, we borrowed that intuition to inform the construction of our model.  We investigate whether the internal latent representations discovered by our models share properties previously identified within linguistics and if not, what important differences exist.  We investigate the interpretability of our model's representations by: 1) A quantitative attachment accuracy and 2) A qualitative look at its output.

Our results corroborate and refute previous work \citep{Hashimoto17,Williams2017}.  We provide stronger evidence that syntactic information can be discovered via latent structured self attention, but we also present preliminary results indicating that conventional definitions of syntax may be at odds with task specific performance.

Unlike in the grammar induction literature our model is not specifically constructed to recover traditional dependency grammars nor have we provided the model with access to part-of-speech tags or universal rules \cite{Naseem:2010aa, Bisk:2013aa}.  The model only uncovers the syntactic information necessary for a given language pair, though future work should investigate if structural linguistic constraints benefit MT.
\subsection{Extracting a Tree}\label{ssec:extract_tree}
For extracting non-projective dependency trees, we use Chu-Liu-Edmonds algorithm \citep{chu-liu-1965,Edmonds}. First, we must collapse BPE segments into words. Assume the $k$-th word corresponds to BPE tokens from index $u$ to $v$. We obtain a new matrix $\hat\mbphi$ by summing over $\mbphi_{i,j}$ that are the corresponding BPE segments\footnote{A visualization of a marginal $\mbbeta$ is given in Appendix~\ref{app:vis}}.
\begin{align}
  \hat\mbphi_{i,j} &=
\begin{cases}
    \mbphi_{i,j} & \text{if } i \not\in [u,v] \land j \not\in [u,v] \\
    \sum_{l=u}^v\mbphi_{i,l}   & \text{if } j=k \land i \not\in [u,v] \\
    \sum_{l=u}^v\mbphi_{l,j} & \text{if }i =k \land j \not\in [u,v] \\
    \sum_{l,h=u}^v\mbphi_{l,h} & \text{otherwise}
  \end{cases}\nonumber
\end{align}

\begin{figure*}[H]
\centering
\begin{tabular}{p{50pt}m{300pt}}
	\textbf{Gold} &
  \scalebox{0.8}{\begin{dependency}[arc edge, arc angle=40, text only label, label style={above}]
    \begin{deptext}
    Its \& ramshackle \& \textit{army} \& had \& virtually \& \textit{collapsed} \& before \& the \& American \& invasion \& in \& 2003 \\
    \end{deptext}
    \depedge{3}{1}{}
    \depedge{3}{2}{}
    \depedge{6}{3}{}
    \depedge[edge style={darkred,thick}]{6}{4}{}
    \depedge{6}{5}{}
    \depedge{10}{7}{}
    \depedge{10}{8}{}
    \depedge{10}{9}{}
    \depedge{6}{10}{}
    \depedge{12}{11}{}
    \depedge{10}{12}{}
  \end{dependency}}\\
  \textbf{SA-NMT}&
  \scalebox{0.8}{\begin{dependency}[arc edge, arc angle=40, text only label, label style={above}]
    \begin{deptext}
    Its \& ramshackle \& \textit{army} \& had \& virtually \& \textit{collapsed} \& before \& the \& American \& invasion \& in \& 2003 \\
    \end{deptext}
    \depedge{10}{8}{}
    \depedge{10}{9}{}
    \depedge{1}{5}{}
    \depedge{3}{1}{}
    \depedge{3}{2}{}
    \depedge[edge style={darkred,thick}]{3}{6}{}
    \depedge{6}{4}{}
    \depedge{7}{3}{}
    \depedge{9}{11}{}
    \depedge{9}{12}{}
    \depedge{12}{7}{}
  \end{dependency}}
  \end{tabular}
\caption{Notice that the model decided \textit{army} is the head of \textit{collapse} because the translation of \textit{collapse} in Russian must agree in number (singular) and gender (feminine) with \textit{army}.}
\label{fig:tree_vis}
\end{figure*}

\subsection{Grammatical Analysis}
\begin{figure*}[ht]
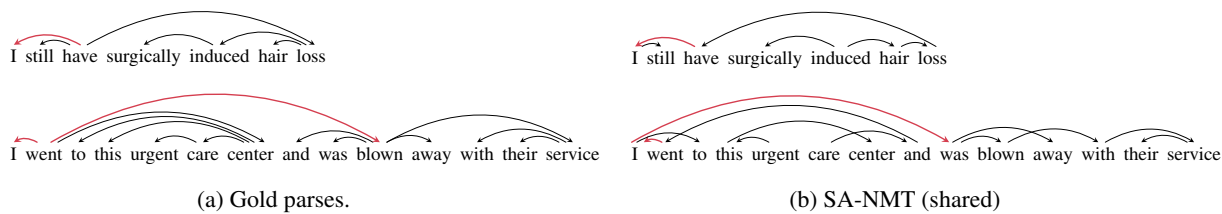

  \begin{subfigure}[b]{0.45\textwidth}
    \scalebox{0.65}{\begin{dependency}[arc edge, arc angle=30, text only label, label style={above},edge slant=0pt]
      \begin{deptext}
      I \& still \& have \& surgically \& induced \& hair \& loss \\
      \end{deptext}
      \depedge[edge style={darkred,thick},edge start x offset=+4pt]{3}{1}{}   
      \depedge[edge start x offset=-2pt]{3}{2}{}  
      \depedge{5}{4}{}  
      \depedge[edge start x offset=+1pt]{7}{5}{}	
      \depedge[edge start x offset=-2pt]{7}{6}{}	
      \depedge[edge end x offset=+2pt]{3}{7}{}  
    \end{dependency}}
    \vfill
    \scalebox{0.65}{\begin{dependency}[arc edge, arc angle=30, text only label, label style={above}]
      \begin{deptext}
      I \& went \& to \& this \& urgent \& care \& center \& and \& was \& blown \& away \& with \& their \& service \\
      \end{deptext}
      \depedge[edge style={darkred,thick}]{2}{1}{}
      \depedge[edge start x offset=+6pt]{7}{3}{}
      \depedge[edge start x offset=+3pt]{7}{4}{}
      \depedge{6}{5}{}
      \depedge{7}{6}{}
      \depedge[edge start x offset=+4pt, edge end x offset=+8pt]{2}{7}{}
      \depedge{10}{8}{}
      \depedge[edge start x offset=-4pt]{10}{9}{}
      \depedge[edge style={darkred,thick}]{2}{10}{}
      \depedge{10}{11}{}
      \depedge{14}{12}{}
      \depedge[edge start x offset=-4pt]{14}{13}{}
      \depedge{10}{14}{}
    \end{dependency}}
    \caption{Gold parses.}
    \label{fig:sample_gold}
  \end{subfigure}\hspace{25pt}
  \begin{subfigure}[b]{0.45\textwidth}
    \scalebox{0.65}{\begin{dependency}[arc edge, arc angle=30, text only label, label style={above}]
      \begin{deptext}
      I \& still \& have \& surgically \& induced \& hair \& loss \\
      \end{deptext}
      \depedge{5}{4}{}
      \depedge{5}{6}{}
      \depedge{1}{2}{}
      \depedge[edge style={darkred,thick}]{3}{1}{}
      \depedge[edge end x offset=-3pt]{6}{7}{}
      \depedge[edge start x offset=+6pt]{7}{3}{}
    \end{dependency}}
    \vfill
    \scalebox{0.65}{\begin{dependency}[arc edge, arc angle=30, text only label, label style={above}]
      \begin{deptext}
      I \& went \& to \& this \& urgent \& care \& center \& and \& was \& blown \& away \& with \& their \& service \\
      \end{deptext}
      \depedge[edge end x offset=-6pt]{6}{8}{}
      \depedge[edge style={darkred,thick},edge start x offset=+2pt, edge end x offset=+5pt]{2}{1}{}
      \depedge[edge start x offset=-3pt]{1}{3}{}
      \depedge[edge style={darkred,thick},edge start x offset=-6pt, edge end x offset=-3pt]{1}{9}{}
      \depedge[edge start x offset=-5pt]{4}{7}{}
      \depedge[edge end x offset=+4pt]{5}{4}{}
      \depedge[edge start x offset=+4pt]{8}{2}{}
      \depedge{9}{10}{}
      \depedge[edge start x offset=-5pt]{9}{11}{}
      \depedge{10}{12}{}
      \depedge{12}{14}{}
      \depedge{14}{13}{}
    \end{dependency}}
    \caption{SA-NMT (shared)}
  	\label{fig:sample_sa_shared}
  \end{subfigure}
  \caption{Samples of induced trees for English by our (En$\to$Ru) model. Notice the red arrows from \textit{subject}$\leftrightarrow$\textit{verb} which are necessary for translating Russian verbs. }
  \label{fig:sample_trees}
\end{figure*}

To analyze performance we compute unlabeled directed and undirected attachment accuracies of our predicted trees on gold annotations from the Universal Dependencies (UD version 2) dataset.\footnote{\url{http://universaldependencies.org}}  We chose this representation because of its availability in many languages, though it is atypical for grammar induction. Our five model settings in addition to left and right branching baselines are presented in Table \ref{DAUA}.  The results indicate that the target language effects the source encoder's induction performance and several settings are competitive with branching baselines for determining headedness.  Recall that syntax is being modeled on the source language so adjacent rows are comparable.

We observe a huge boost in DA/UA scores for EN and RU in FA-NMT and SA-NMT-shared models when the target languages are morphologically rich (RU and AR respectively). In comparison to previous work \citep{Belinkov17,shi:2016} on an encoder's ability to capture source side syntax, we show a stronger result that even when the encoders are designed to capture syntax explicitly, the choice of the target language influences the amount of syntax learned by the encoder.

We also see gains from  hard attention and several models outperform baselines for undirected dependency metrics (UA). Whether hard attention helps in general is unclear.  It appears to help when the target languages are morphologically rich.

Successfully extracting linguistic structure with hard attention indicates that models can capture interesting structures beyond semantic co-occurrence via discrete actions.
This corroborates previous work \citep{Choi2017,Yogatama16} which has shown that non-trivial structures are learned by using REINFORCE \citep{williams:1992} or the Gumbel-softmax trick \citep{jang2016} to backprop through discrete units.
Our approach also outperforms \cite{Hashimoto17} despite lacking access to additional resources like POS tags.\footnote{The numbers are not directly comparable since they use WSJ corpus to evaluate the UA score.}

\subsection{Dependency Accuracies \& Discrepancies}
While the SA-NMT-hard model gives the best directed attachment scores on EN$\to$DE, DE$\to$EN and RU$\to$AR, the BLEU scores of this model are below other SA-NMT models as shown in Table~\ref{tb:bleus}. The lack of correlation between syntactic performance and NMT contradicts the intuition of previous work and suggests that useful structures learned in service of a task might not necessarily benefit from or correspond directly to known linguistic formalisms. We want to raise three important differences between these induced structures and UD.

First, we see a blurred boundary between dependency and constituency representations.  As noted earlier, the BiLSTM provides a local summary.
When the model chooses a head word, it is actually choosing hidden states from a BiLSTM and therefore gaining access to a constituent or region.  This means there is likely little difference between attending to the noun vs the determiner in a phrase (despite being wrong according to UD).  Future work might force this distinction by replacing the BiLSTM with a bag-of-words but this will likely lead to substantial losses in MT performance.

Second, because the model appears to use syntax for agreement, often verb dependencies link to subjects directly to capture predicate argument structures like those in CCG or semantic role labeling.  UD instead follows the convention of attaching all verbs that share a subject to one another or their conjunctions.  We have colored some subject--verb links in Figure \ref{fig:sample_trees}: {\em e.g.}, between \textit{I}, \textit{went} and \textit{was}.

Finally, the model's notion of headedness is atypical as it roughly translates to ``helpful when translating''.  The head word gets incorporated into the shared representation which may cause the arrow to flip from traditional formalisms.
Additionally, because the model can turn on and off syntax as necessary, it is likely to produce high confidence treelets rather than complete parses.  This means arcs produced from words with weak gate activations (Figure \ref{fig:gatenorm}) are not actually used during translation and likely not-syntactically meaningful.

We will not speculate if these are desirable properties or issues to address with constraints, but the model's decisions appear well motivated and our formulation allows us to have the discussion.

\section{Conclusion}
We have proposed a structured self attention encoder for NMT. Our models show significant gains in performance over a strong baseline on standard WMT benchmarks. The models presented here do not access any external information such as parse-trees or part-of-speech tags yet appear to use and induce structure when given the opportunity.  Finally, we see our induction performance is language pair dependent, which invites an interesting research discussion as to the role of syntax in translation and the importance of working with morphologically rich languages.

\section*{Acknowledgments}
We thank Joachim Daiber, Ekaterina Garmash, and Julia Kiseleva for helping with German and Russian examples.  We are grateful to Arianna Bisazza, Milo\v s Stanojevi\'c, and Raquel Garrido Alhama for providing feedback on the draft.
The second author was supported by Samsung Research.

\bibliography{biblio}
\bibliographystyle{acl_natbib}
 \appendix

 \section{Attention Visualization}
 \label{app:vis}
\begin{figure}[ht]
   \centering
   \begin{subfigure}[b]{0.45\textwidth}
     \includegraphics[width=\textwidth]{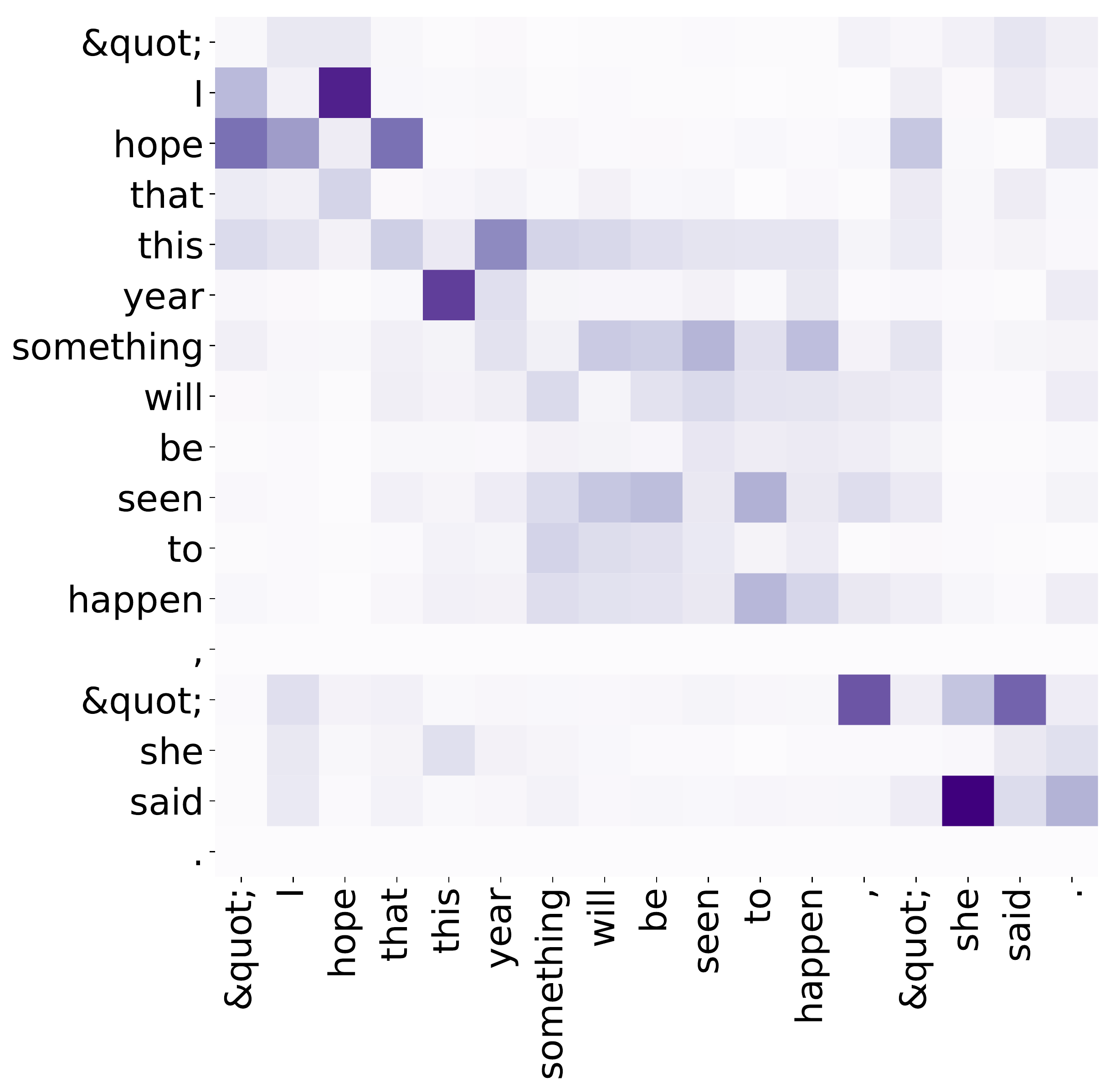}
     \caption{SA-NMT (shared) attention.}
     \label{fig:soft_attn}
   \end{subfigure}
   \quad
   \begin{subfigure}[b]{0.45\textwidth}
     \includegraphics[width=\textwidth]{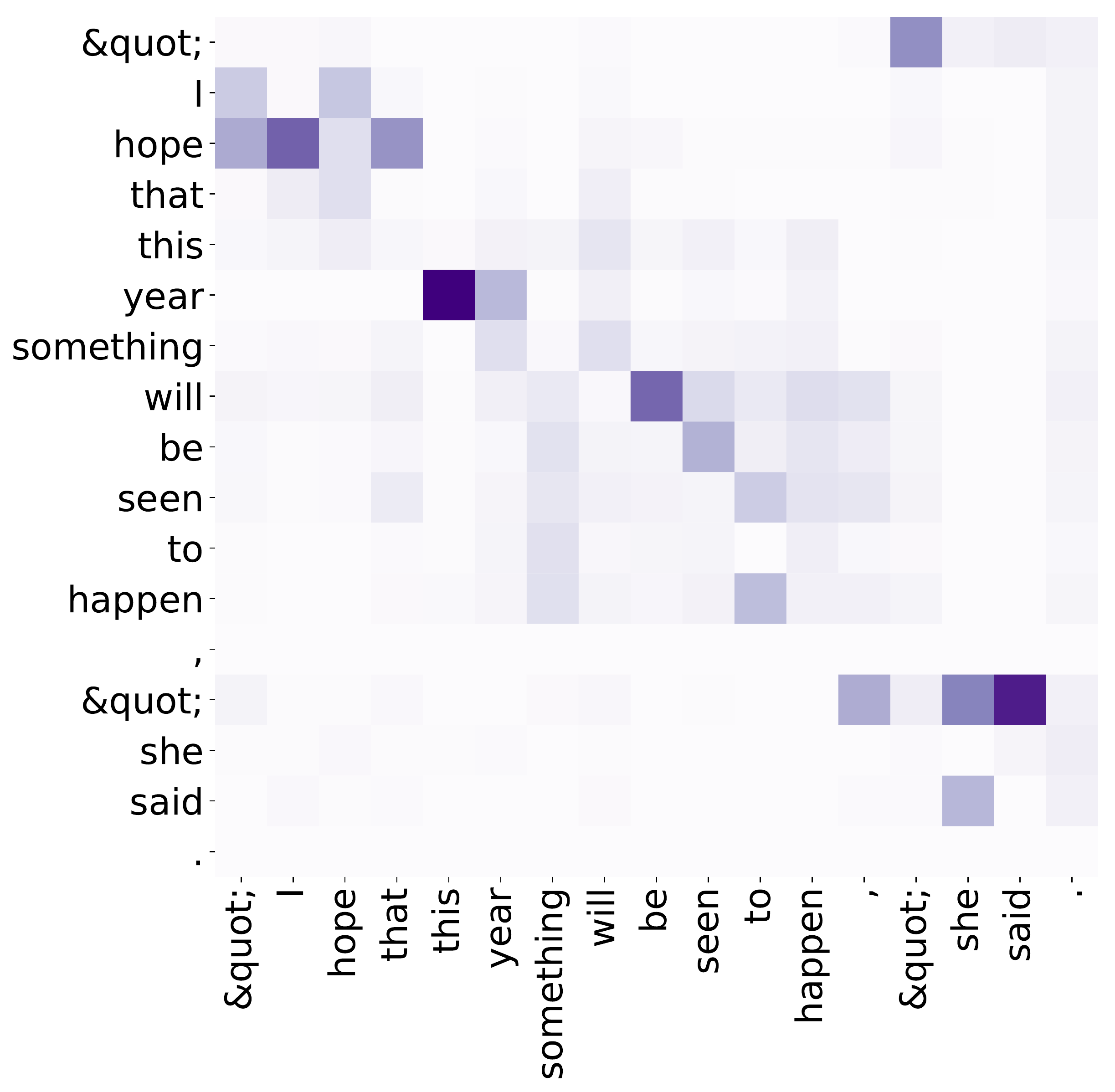}
     \caption{SA-NMT with hard structured attention.}
     \label{fig:hard_attn}
   \end{subfigure}
   \caption{A visualization of attention distributions over head words (on y-axis).}\label{fig:vis_sa_attn}
 \end{figure}

 Figure~\ref{fig:vis_sa_attn} shows a sample visualization of structured attention models trained on En$\to$De data.
 It is worth noting that the shared SA-NMT model (Figure~\ref{fig:soft_attn}) and the hard SA-NMT model (Figure~\ref{fig:hard_attn}) capture similar structures of the source sentence. We hypothesize that when the objective function requires syntax, the induced trees are more consistent unlike those discovered by a semantic objective \citep{Williams2017}. Both models correctly identify that the verb is the head of pronoun (hope$\to$I, said$\to$she). While intuitively it is clearly beneficial to know the subject of the verb when translating from English into German, the model attention is still somewhat surprising because long distance dependency phenomena are less common in English, so we would expect that a simple content based addressing ({\em i.e.} standard attention mechanism) would be sufficient in this translation
\end{document}